\documentclass{article}

\usepackage{PRIMEarxiv}

\usepackage[utf8]{inputenc} 
\usepackage[T1]{fontenc}    
\usepackage{hyperref}       
\usepackage{url}            
\usepackage{booktabs}       
\usepackage{amsfonts}       
\usepackage{nicefrac}       
\usepackage{microtype}      
\usepackage{lipsum}
\usepackage{fancyhdr}       
\usepackage{graphicx}       
\graphicspath{{media/}}     

\title{Myopia prediction for adolescents via time-aware deep learning
}

\author{
  Junjia Huang \\
  Computational Education Lab, Big Data Research Center \\
  University of Electronic Science and Technology of China \\
  Chengdu, China\\
  \texttt{coderjia@std.uestc.edu.cn } \\
   \And
  Wei Ma* \\
  Department of Ophthalmology of West China Hospital \\
  Sichuan University \\
  Chengdu, China\\
  \texttt{mawei@wchscu.cn} \\
   \AND
   Rong Li \\
   Eye See Inc. \\
   Chengdu, China\\
   \And
   Na Zhao \\
   National Pilot School of Software \\
   Yunnan University \\
   Kunming, China\\
   \And
   Tao Zhou*\\
   Computational Education Lab, Big Data Research Center \\
   University of Electronic Science and Technology of China \\
   Chengdu, China\\
   \texttt{zhutou@ustc.edu} \\
}

\usepackage{multirow} 
\usepackage{threeparttable}
\begin{document}
\maketitle

\begin{abstract}
Background: Quantitative prediction of the adolescents’ spherical equivalent based on their variable-length historical vision records.

Methods: From October 2019 to March 2022, we examined binocular uncorrected visual acuity, axial length, corneal curvature, and axial of 75,172 eyes from 37,586 adolescents aged 6-20 years in Chengdu, China. 80\% samples consist of the training set and the remaining 20\% form the testing set. Time-Aware Long Short-Term Memory was used to quantitatively predict the adolescents’ spherical equivalent within two and a half years.

Result: The mean absolute prediction error on the testing set was 0.273 ± 0.257 for spherical equivalent, ranging from 0.189 ± 0.160 to 0.596 ± 0.473 if we consider different lengths of historical records and different prediction durations. 

Conclusions: Time-Aware Long Short-Term Memory was applied to captured the temporal features in irregularly sampled time series, which is more in line with the characteristics of real data and thus has higher applicability, and helps to identify the progression of myopia earlier. The overall error 0.273 is much smaller than the criterion for clinically acceptable prediction, say 0.75.
\end{abstract}


\section{Background}
Myopia is a global public health concern. It is estimated that 57\% of countries will have a myopia prevalence of more than 50\% by 2050 \cite{Holden2016}. The World Health Organization reported in 2019 that at least 2.2 billion people have a vision impairment, of whom at least 1 billion have a vision impairment that could have been prevented \cite{who2019}. As myopia is currently difficult to be cured completely, it is vital to prevent its onset and progression. An early and appropriate intervention can effectively mitigate the risks and consequences related to myopia \cite{Sankaridurg2021}. The spherical equivalent (SE) is the basis for screening and diagnosing myopia \cite{Jong2022}. Quantitative prediction of SE can indicate the specific changes in the progression of myopia, and help in designing targeted interventions in advance. Previous studies have reported a number of risk factors for the onset or progression of myopia, including age, gender, heredity, outdoor activities, etc. \cite{French2013,Fan2016,Tedja2019}. Matsumura \textit{et al.} suggested that the historical progression of myopia is associated with future changes in visual acuity \cite{Matsumura2020}. Therefore, we believe that the historical vision records, together with other demographic information, can be used to quantitatively predict SE.

In recent years, a growing body of research has considered the prediction of myopia or high myopia in different populations \cite{Han2022}. Most known studies used traditional models like linear regression, support vector machines, decision trees, and so on \cite{Jones2007,Chua2016,ngm2018,Lin2018,Ma2018,Chen2019,Yang2020,Huang2021}. In comparison, deep learning can be trained with complex and nonlinear parameters to learn data structures \cite{VARGAS2018}, and is deemed to perform better than traditional models in a variety of medical prediction tasks \cite{Siuly2016,Adam2020,Jalali2020,Sun2021}. However, there are only few applications of deep learning in myopia prediction. 

Spadon \textit{et al.} argued that the temporal dynamics provides valuable information in addition to static symptom observation \cite{Spadon2021}. However, in usual vision records, the uneven distribution of time intervals between historical records makes the extraction of temporal features very difficult. This paper uses Time-Aware Long Short-Term Memory (T-LSTM) to capture the temporal features in irregularly sampled time series, and to quantitatively predict adolescents’ SE based on their variable-length historical vision records. The proposed method is widely applicable. 

\section{Methods}
\subsection{Data Description}
The dataset for this study contains 232,244 historical vision records from 37,586 school-aged adolescents (aged 6-20 years) in Chengdu, China. They were collected by Eye See Inc. from October 2019 to March 2022 through unscheduled refractive screening in schools. Inclusion criteria: elementary, middle and high school students between the ages of 6 and 20. Exclusion criteria: students who did not obtain consent from their parents or their guardians, students who were unable to cooperate with the examination or did not complete the examination due to intellectual or physical reasons. The following examinations were performed based on the standard clinical protocols: (1) distant vision examination; (2) slit-lamp microscope examination; (3) refractive examination; (4) axial eye length measurement. 

The myopia diagnostic criteria associated were developed in accordance with the Consensus on Myopia Management for Asia 2021, published by the Asia Optometric Management Academy (AOMA) and Asia Optometric Congress (AOC) \cite{Jong2022}. Based on SE when the eye is relaxed, the criterion of myopia is \(SE \leq -0.5 D\), and the degree of myopia is classified as follows: (1) low myopia: \(-3.0 D < SE \leq -0.5 D\); (2) moderate myopia: \(-6.0 D < SE \leq -3.0 D\); (3) high myopia: \(SE \leq -6.0 D\).

\begin{figure}[!h]
    \caption{The distributions of the 14 features.}
    \centering
    \includegraphics[width=16cm]{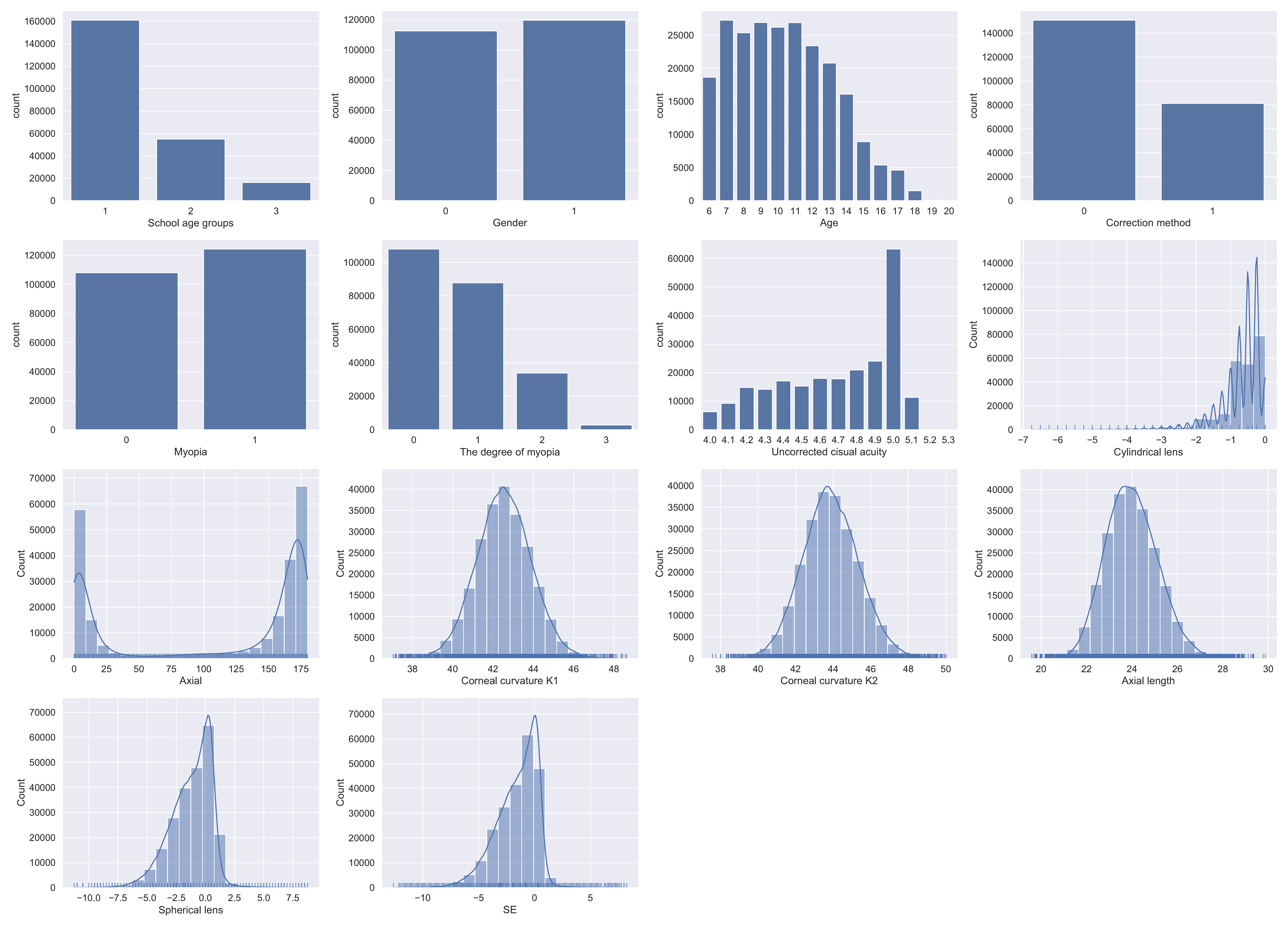}
    \label{fig:figure1}
\end{figure}

The cleaned dataset contains 75,172 eyes (samples) of 37,586 adolescents. Each sample is associated with 2 to 6 records. The number of samples with 2, 3, 4, 5 and 6 records is 27,015, 18,732, 25,109, 4,314 and 2, respectively.  The interval time between the first record and the last record for any sample ranges from 1 (\(< 1 quarter\)) to 10 (\(\geq 9 quarters, < 10 quarters\)). Each record is associated with 16 features, as described in Table \ref{tab:table1}. Figure \ref{fig:figure1} shows distributions of the 14 features. 

\begin{table}[!h]
    \caption{Feature description. Discrete variables: value. Continuous variables: mean ± standard deviation, {[}min, max{]}.}
    \centering
    \label{tab:table1}
    \begin{tabular}{cc}
    \hline
    \textbf{Features}         & \textbf{\begin{tabular}[c]{@{}c@{}}Statistics \end{tabular}} \\ \hline
    Id                        &                                                                                                                                                            \\
    Check date                &                                                                                                                                                            \\
    School age groups         & 1: elementary school, 2: middle school, 3: high school                                                                                                     \\
    Gender                    & 0: female, 1: male                                                                                                                                         \\
    Age                       & 10.38 ± 2.90, {[}6, 20{]}                                                                                                                                  \\
    Correction method         & 0: uncorrected, 1: spectacles glasses                                                                                                                      \\
    Uncorrected visual acuity & 4.69 ± 0.32, {[}4.0, 5.3{]}                                                                                                                                \\
    Spherical lens            & -1.21 ± 1.75, {[}-11.25, 8.75{]}                                                                                                                           \\
    Cylindrical lens          & -0.71 ± 0.65, {[}-6.75, 0.00{]}                                                                                                                            \\
    Axial                     & 106.52 ± 76.12, {[}0,180{]}                                                                                                                                \\
    Corneal curvature K1      & 42.54 ± 1.36, {[}37.05, 48.63{]}                                                                                                                           \\
    Corneal curvature K2      & 43.87 ± 1.51, {[}37.58, 50.00{]}                                                                                                                           \\
    Axial length              & 23.99 ± 1.14, {[}18.59, 29.86{]}                                                                                                                           \\
    Myopia                    & 0: no myopia, 1: myopia                                                                                                                                    \\
    The degree of myopia      & 0: no myopia, 1: low myopia, 2: moderate myopia, 3: high myopia                                                                                            \\
    SE                        & -1.57 ± 1.85, {[}-12.63, 8.25{]}                                                                                                                           \\ \hline
    \end{tabular}
\end{table}

\subsection{Data Preprocessing}
Firstly, in order to exclude the interference between categories of the original sequential encoding, one-hot encoding was performed for the unordered categorical features, say correction method and gender. It creates unit vectors for each option within the categorical feature, where the dimensionality of the vector equals the number of categories \cite{Satya2021}. For example, a possible one-hot encoding for gender, as above, is male: (1, 0), and female: (0, 1). 

After one-hot encoding, the features were standardized except for Id and Check date to speed up the convergence of the model. The standardization rescales the sample mean to zero (\(\mu=0\)) and variance to unit (\(\sigma=1\)) \cite{Sharma2018}, as
\begin{equation}
    x^\prime=\frac{x\ -\ \mu}{\sigma}.
\end{equation}
To increase the sample size, the historical records of an adolescent were split into several samples, ensuring that all input data used for training and predicting is recorded before the label (i.e., the SE value). For example, an adolescent’s 4 records (a, b, c, d) can be split into 11 samples as shown in Table \ref{tab:table2}.

\begin{table}[!h]
    \caption{The enhanced samples from an adolescent’s 4 records (a, b, c, d). The time interval [ab] means the interval between record a and record b, in quarter.}
    \centering
    \label{tab:table2}
    \begin{tabular}{ccc}
    \hline
    \textbf{Input Data} & \textbf{Label} & \textbf{Time Interval} \\ \hline
    {[}a{]}             & {[}b{]}        & {[}ab{]}               \\
    {[}a{]}             & {[}c{]}        & {[}ac{]}               \\
    {[}a{]}             & {[}d{]}        & {[}ad{]}               \\
    {[}b{]}             & {[}c{]}        & {[}bc{]}               \\
    {[}b{]}             & {[}d{]}        & {[}bd{]}               \\
    {[}c{]}             & {[}d{]}        & {[}cd{]}               \\
    {[}a, b{]}          & {[}c{]}        & {[}ab, bc{]}           \\
    {[}a, b{]}          & {[}d{]}        & {[}ab, bd{]}           \\
    {[}a, c{]}          & {[}d{]}        & {[}ac, cd{]}           \\
    {[}b, c{]}          & {[}d{]}        & {[}bc, cd{]}           \\
    {[}a, b, c{]}       & {[}d{]}        & {[}ab, bc, cd{]}       \\ \hline
    \end{tabular}
\end{table}

The number of samples in the dataset increased to 490,420 after the data preprocessing. The sample sizes are 277,035, 162,348, 46,709, 4,326 and 2 for sequence lengths of 1, 2, 3, 4 and 5, respectively. Particularly, sample with sequence length 5 is too few to be included in the training. The dataset was then divided into layers by the lengths of sequences. Each layer was further divided into training set (80\%) and testing set (20\%). 

\subsection{LSTM}
Recurrent Neural Network (RNN) is a neural network structure that can effectively link contextual information to achieve long term memory, but suffer from the problem of gradient vanishing or exploding \cite{Pascanu2013,Zaremba2014}. To solve this challenge, Hochreiter \textit{et al.} proposed the method named Long Short-Term Memory (LSTM) \cite{Hochreiter1997}, which is a variant of RNN, combining short-term memory with long-term memory through gate control. LSTM solves the problem of gradient vanishing to a certain extent and allows for the learning of long-term dependent information.

Standard LSTM unit (Figure \ref{fig:figure2}(a)) consists of a forget gate, an input gate, an output gate and a cell state. The current state \(h_t\) is influenced by the previous state \(h_{t-1}\) and the current input \(x_t\). 

Forget gate:

\begin{equation}
    f_t=\sigma\left(W_fx_t+U_fh_{t-1}+b_f\right)
\end{equation}

Input gate:

\begin{equation}
    i_t=\sigma\left(W_ix_t+U_ih_{t-1}+b_i\right)
\end{equation}
\begin{equation}
    \widetilde{C_t}=tanh{\left(W_cx_t+U_ch_{t-1}+b_c\right)}
\end{equation}

Output gate:

\begin{equation}
    o_t=\sigma\left(W_ox_t+U_oh_{t-1}+b_o\right)
\end{equation}
\begin{equation}
    h_t=o_t\cdot t a n h{\left(C_t\right)}
\end{equation}

Cell state:

\begin{equation}
    C_t=f_tC_{t-1}+i_t\widetilde{C_t}
\end{equation}

where \(\sigma\) and \(tanh\) represent the activation functions, and \(W, U\) and \(b\) are the learnable parameters.

The standard LSTM assumes that the time intervals between sequential elements are uniformly distributed, and thus cannot handle the problem with irregular time intervals. 

\subsection{T-LSTM}
T-LSTM (Figure \ref{fig:figure2}(b)) introduces time interval information based on the standard LSTM, and attenuates the short-term memory according to the time intervals in order to capture the temporal dynamics of the sequential data with temporal irregularity \cite{Baytas2017}. T-LSTM accepts two inputs: the current record and the current time step elapsed. T-LSTM differs from the standard LSTM primarily in the subspace decomposition of the previous time step, which adjusts the short-term memory according to the time intervals between records. The subspace decomposition method does not change the effect of the current input on the current output, but changes the effect of the previous memory on the current output. Specifically, T-LSTM adds the following features to the standard LSTM: (i) Short-term memory \(C_{t-1}^S\), obtained through the memory of the previous time step, as 

\begin{equation}
    C_{t-1}^S=tanh{\left(W_dC_{t-1}+b_d\right)}.
\end{equation}

(ii) Discounted short-term memory \({\hat{C}}_{t-1}^S\), obtained by weighting \(C_{t-1}^S\) with time elapsed, as

\begin{equation}
    {\hat{C}}_{t-1}^S=C_{t-1}^S\cdot g\left(\Delta_t\right).
\end{equation}

(iii) Long-term memory \(C_{t-1}^T\), which is the supplementary subspace of short-term memory, as

\begin{equation}
    C_{t-1}^T=C_{t-1}-C_{t-1}^S.
\end{equation}

(iv) Adjusted previous memory \(C_{t-1}^\ast\), obtained through combining discounted short-term memory and long-term memory, as

\begin{equation}
    C_{t-1}^\ast=C_{t-1}^T+{\hat{C}}_{t-1}^S.
\end{equation}

\subsection{Application of T-LSTM in myopia prediction}
\begin{figure}[!h]
    \caption{The structure of LSTM, T-LSTM and T-LSTM in myopia prediction, where \(x\) denotes the temporal input data, \(C\) is the cell state representing the long-term memory, \(h\) is the hidden state representing the short-term memory, \(\mathrm{\Delta}_t\) is the time interval between records \(x_t\) and \(x_{t-1}\), \(\sigma\) is the sigmoid activation function, and \(tanh\) is the tanh activation function.}
    \centering
    \includegraphics[width=18cm]{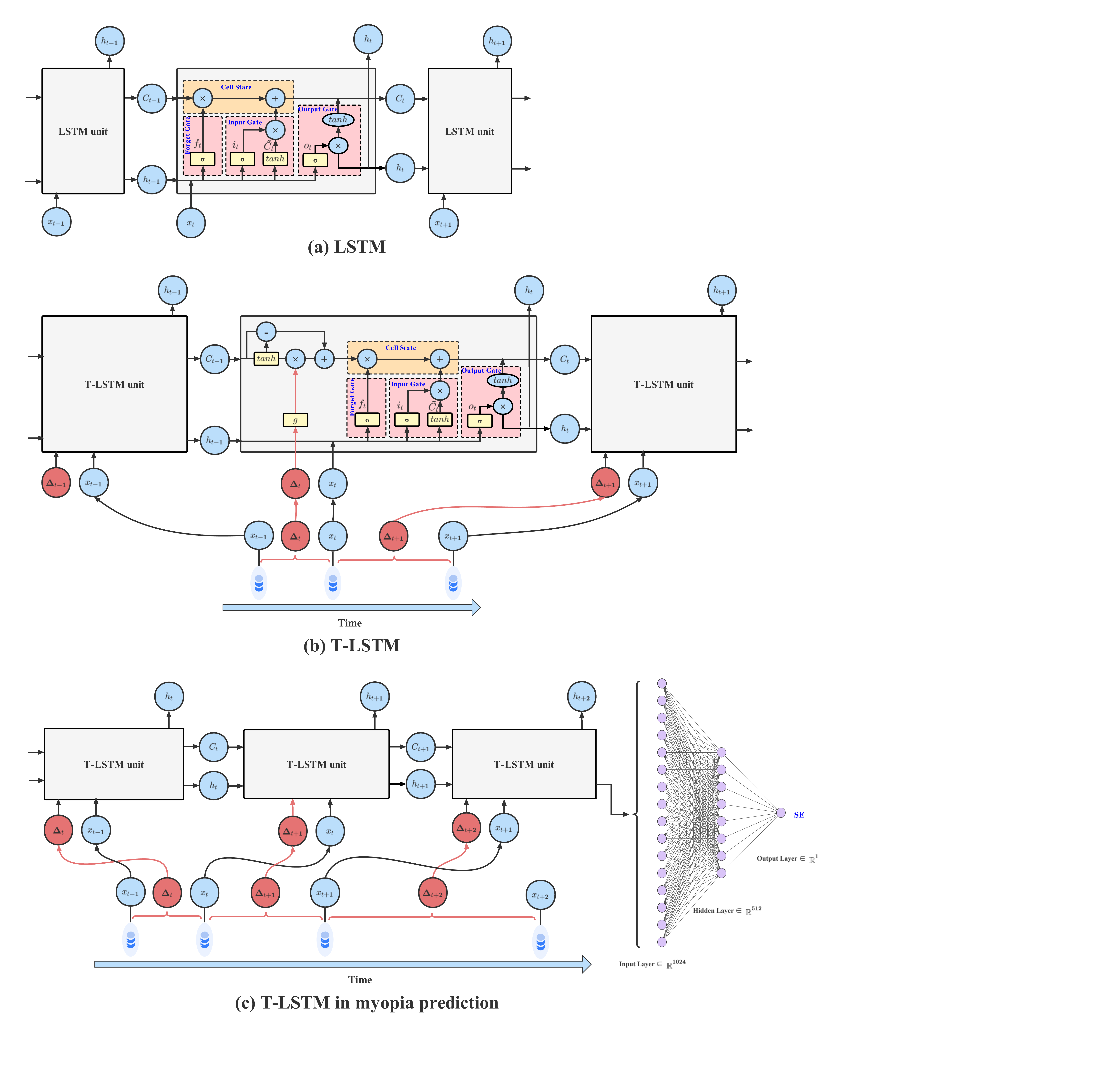}
    \label{fig:figure2}
\end{figure}

The input of each cell of T-LSTM is the current record \(x_t\) and the time interval \(\mathrm{\Delta}_t\) between \(x_{t-1}\) and \(x_t\). The output is the current state \(h_t\). In the myopia prediction model proposed in this paper, the input of each cell is changed to the current record \(x_t\) and the time interval \(\mathrm{\Delta}_{t+1}\) between \(x_t\) and \(x_{t+1}\). The output is the next state \(h_{t+1}\). The final prediction is the output of the last step which is passed through the fully connected neural network. The structure of the model is shown in Figure \ref{fig:figure2}(c). When performing myopia prediction, the values of visual acuity at any future moment can be predicted by changing the value of the last time interval. 

\subsection{Metrics}
The model’s objective function is the mean square error (MSE) of SE, often referred to as the loss. The MSE lies in the range \([0, +\infty)\), as

\begin{equation}
    MSE=\frac{1}{m}\sum_{i=1}^{m}\left(y_i-{\hat{y}}_i\right)^2,
    \label{eq:eq12}
\end{equation}

where \(y_i\) is the actual value, \({\hat{y}}_i\) is the predicted value, and \(m\) is the number of samples. Eq. (\ref{eq:eq12}) is a smooth, continuous and everywhere derivable function, and thus being convenient for the gradient descent algorithm.
The prediction performance of the model is evaluated by the mean absolute error (MAE), which is the average of the absolute deviations, as
\begin{equation}
    MAE=\frac{1}{m}\sum_{i=1}^{m}\left|\left(y_i-{\hat{y}}_i\right)\right|.
\end{equation}

It takes values in the range of \([0, +\infty)\). A smaller MAE indicates a better model. 

\section{Results}
\begin{figure}[!h]
    \caption{The change of MSE in the model. An epoch means training the neural network with all the training data for one cycle.}
    \centering
    \includegraphics[width=10cm]{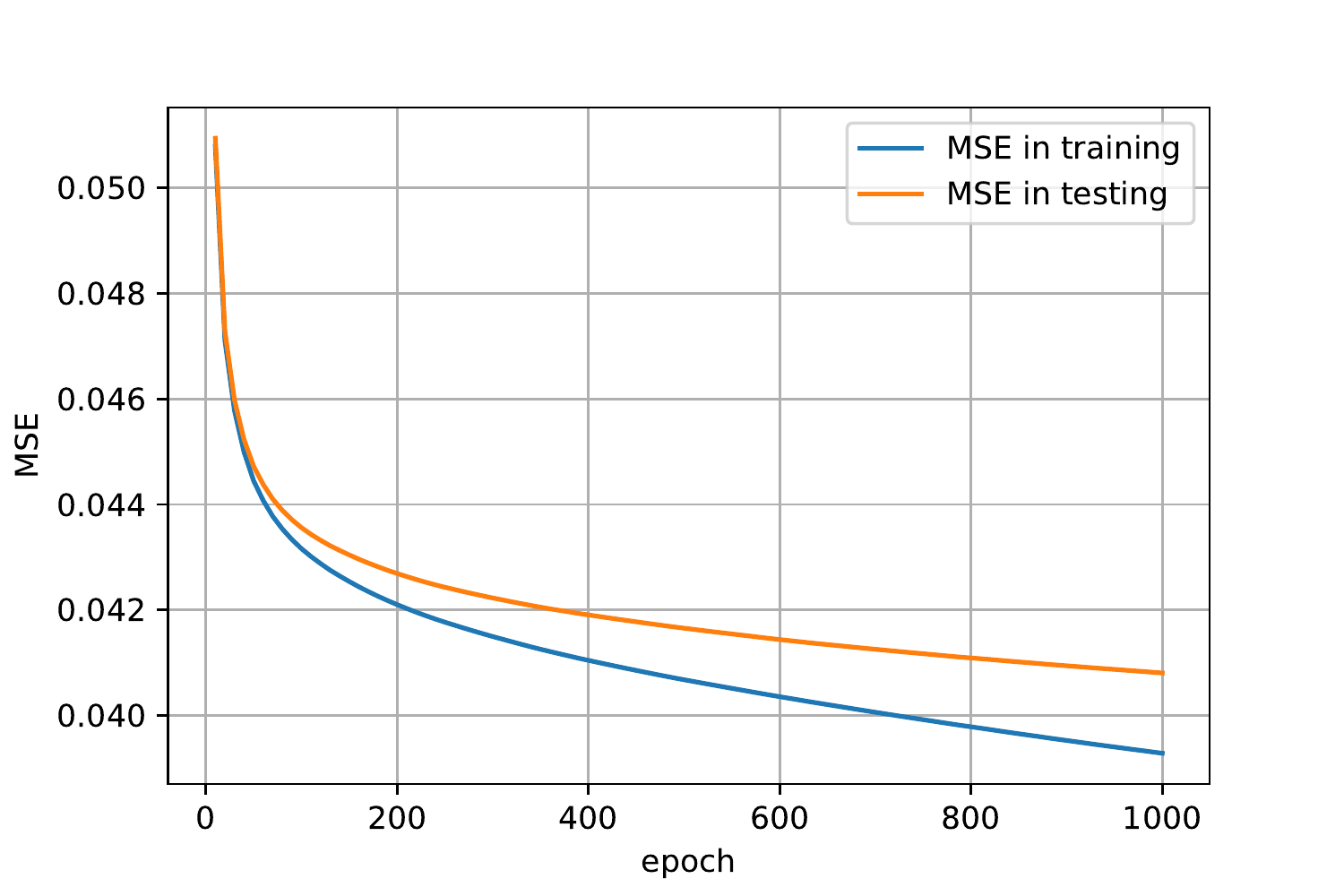}
    \label{fig:figure3}
\end{figure}

After 1000 training iterations, the model converges with the loss (i.e., MSE) of the training process displayed in Figure \ref{fig:figure3}. The MAE of future SE is 0.273 ± 0.257 on the testing set. The stratified MAE is shown in Table \ref{tab:table3}. When sequence lengths are 1, 2, 3 and 4, the corresponding MAE ranges from 0.227 to 0.596 for 2 to 10 quarters, 0.205 to 0.380 for 2 to 6 quarters, 0.201 to 0.239 for 2 to 4 quarters and 0.189 for 2 quarters, respectively. Overall speaking, the longer the sequence length and the shorter the prediction duration, the smaller the prediction error. The MAE of SE within 0.75 is considered to be a clinically acceptable prediction \cite{Lin2018}. Based on the accuracy and robustness of the model, as well as the variance of the prediction performance, the model provides a clinically valuable prediction of adolescents’ vision in the short and medium term. 

\begin{table}[!h]
    \caption{The MAE of SE in the testing set.}
    \centering
    \label{tab:table3}
    \begin{threeparttable}
        \begin{tabular}{ccccc}
        \hline
        \multirow{2}{*}{\textbf{Prediction duration\tnote{a}}} & \multicolumn{4}{c}{\textbf{\begin{tabular}[c]{@{}c@{}}The MAE of SE for different sequence lengths \\ (mean ± standard deviation (sample size))\tnote{b}\end{tabular}}} \\ \cline{2-5} 
                                                      & \textbf{1}                               & \textbf{2}                               & \textbf{3}                             & \textbf{4}                            \\ \hline
        \textbf{1}                                    & *0.447 ± 0.503 (13)                        & *0.261 ± 0.144 (17)                        & *0.086 ± 0.079 (4)                       &                                       \\
        \textbf{2}                                    & 0.227 ± 0.224 (9401)                       & 0.205 ± 0.203 (12832)                      & 0.201 ± 0.182 (7318)                     & 0.189 ± 0.160 (865)                     \\
        \textbf{3}                                    & 0.247 ± 0.208 (18678)                      & 0.229 ± 0.180 (10531)                      & 0.230 ± 0.171 (1098)                     & *0.021 ± 0.000 (1)                      \\
        \textbf{4}                                    & 0.304 ± 0.282 (8198)                       & 0.278 ± 0.252 (6800)                       & 0.239 ± 0.198 (922)                      &                                       \\
        \textbf{5}                                    & 0.337 ± 0.274 (8732)                       & 0.316 ± 0.240 (1282)                       &                                        &                                       \\
        \textbf{6}                                    & 0.414 ± 0.350 (7151)                       & 0.380 ± 0.296 (1007)                       &                                        &                                       \\
        \textbf{7}                                    & 0.475 ± 0.419 (618)                        & *0.397 ± 0.000 (1)                         &                                        &                                       \\
        \textbf{8}                                    & 0.485 ± 0.340 (321)                        &                                          &                                        &                                       \\
        \textbf{9}                                    & 0.550 ± 0.412 (1150)                       &                                          &                                        &                                       \\
        \textbf{10}                                   & 0.596 ± 0.473 (1145)                       &                                          &                                        &                                       \\ \hline
        \end{tabular}
        \begin{tablenotes}
            \footnotesize
            \item a. a value \(i\) means the duration ranges from \(i-1\) quarters to \(n\) quarters.
            \item b. * represents that the sample size is too small (<100) to be a solid reference.
        \end{tablenotes}
    \end{threeparttable}
\end{table}

\section{Conclusions}
As the symptoms of myopia are not typical, they are often ignored by parents in the early stages of development. However, if low myopia is not controlled, it can lead to high myopia and very serious blinding ocular complications, such as posterior scleral and macular degeneration, as well as a substantially higher chance of developing cataracts and glaucoma \cite{Saw2005,Neelam2012}. The earlier the onset of myopia, the more likely the eye axial length will elongate, the faster myopia will progress, and the higher the final diopter \cite{Hu2020}. This paper can quantitatively predict the adolescents’ SE within two and a half years, and help to identify the progression of myopia earlier so that targeted interventions and corrective measures can be taken. This is of great significance for the prevention and control of myopia. 

As the development of myopia is affected by a number of complex factors, such as heredity, environment, and behaviors \cite{Guo2016,Choi2017}, to achieve accurate myopia predictions is challenging. Deep learning is able to infer new features from the limited sets of features contained in the training set, while avoiding complex feature engineering. This paper applied T-LSTM to captured the temporal features in irregularly sampled time series, which is more in line with the characteristics of real data and thus has higher applicability.

\section{Discussion}
To the best of our knowledge, only a very small number of studies include quantitative predictions of future visual acuity. Among them, Lin \textit{et al.} achieved quantitative prediction of future SE in a study of nearly 130,000 people in Guangdong, China, 2018, where the MAE for 1 to 8-year SE prediction ranges from 0.253 to 0.799 \cite{Lin2018}. This paper achieves close prediction accuracy on a smaller dataset. This study can indicate the trend of refraction and visual acuity in the next two and a half years. The results are interesting not only for medical institutions to make statistics, but also for parents to see the degree of vision loss more intuitively. In this way, it will guide guardians to take their children for timely myopia correction and early myopia prevention and control, which is more important and proactive than the post intervention by medical institutions and will contribute to the prevention and control of early myopia in adolescents. 

The current study has some limitations. Firstly, the sample area is concentrated, and thus the representation is insufficient. Secondly, the depth of longitudinal data still needs to be enhanced. Thirdly, the model only predicts visual acuity based on historical vision records and does not take into account the effects of genetic and environmental factors, which could be analyzed in the future with the help of family survey.

\bibliographystyle{elsarticle-num}
\bibliography{myreference}

\end{document}